\pdfoutput=1

\documentclass[11pt]{article}

\usepackage{emnlp2021}

\usepackage{times}
\usepackage{latexsym}
\usepackage{amsmath}
\usepackage{graphicx}
\usepackage{algorithm}
\usepackage{algorithmic}
\usepackage[T1]{fontenc}

\usepackage[utf8]{inputenc}

\usepackage{paralist}

\usepackage{microtype}
\usepackage{multirow}
\title{Guiding Topic Flows in the Generative Chatbot by Enhancing the ConceptNet with the Conversation Corpora}

\author{Pengda Si \\
    Tsinghua University, Shenzhen, China \\
    \texttt{spd18@tsinghua.org.cn} \\ \And
    Yao Qiu\\
    Tencent Inc, Beijing, China \\
    \texttt{yasinqiu@tencent.com} \\ \AND
    Jinchao Zhang\\
    Tencent Inc, Beijing, China \\
    \texttt{jinchaozhang@tencent.com} \\ \And
    Yiru Wang \\
    Tencent Inc, Shenzhen, China \\
    \texttt{wangyiru017@gmail.com} \\ \AND
    Jie Zhou \\
    Tencent Inc, Beijing, China \\
    \texttt{withtomzhou@tencent.com} \\ \And
    Yujiu Yang \\
    Tsinghua University, Shenzhen, China \\
    \texttt{yang.yujiu@sz.tsinghua.edu.cn} \\
}

\begin{document}

\maketitle
\begin{abstract}
Human conversations consist of reasonable and natural topic flows, which are observed as the shifts of the mentioned concepts across utterances.
Previous chatbots that incorporate the external commonsense knowledge graph prove that modeling the concept shifts can effectively alleviate the dull and uninformative response dilemma.
However, there still exists a gap between the concept relations in the natural conversation and those in the external commonsense knowledge graph, which is an issue to solve.
Specifically, the concept relations in the external commonsense knowledge graph are not intuitively built from the conversational scenario but the world knowledge, which makes them insufficient for the chatbot construction.
To bridge the above gap, we propose the method to supply more concept relations extracted from the conversational corpora and reconstruct an enhanced concept graph for the chatbot construction.
In addition, we present a novel, powerful, and fast graph encoding architecture named the Edge-Transformer to replace the traditional GNN architecture.
Experimental results on the Reddit conversation dataset indicate our proposed method significantly outperforms strong baseline systems and achieves new SOTA results.
Further analysis individually proves the effectiveness of the enhanced concept graph and the Edge-Transformer architecture.

\end{abstract}

\section{Introduction}


\begin{figure}[ht]
\centering
\includegraphics[width=\columnwidth]{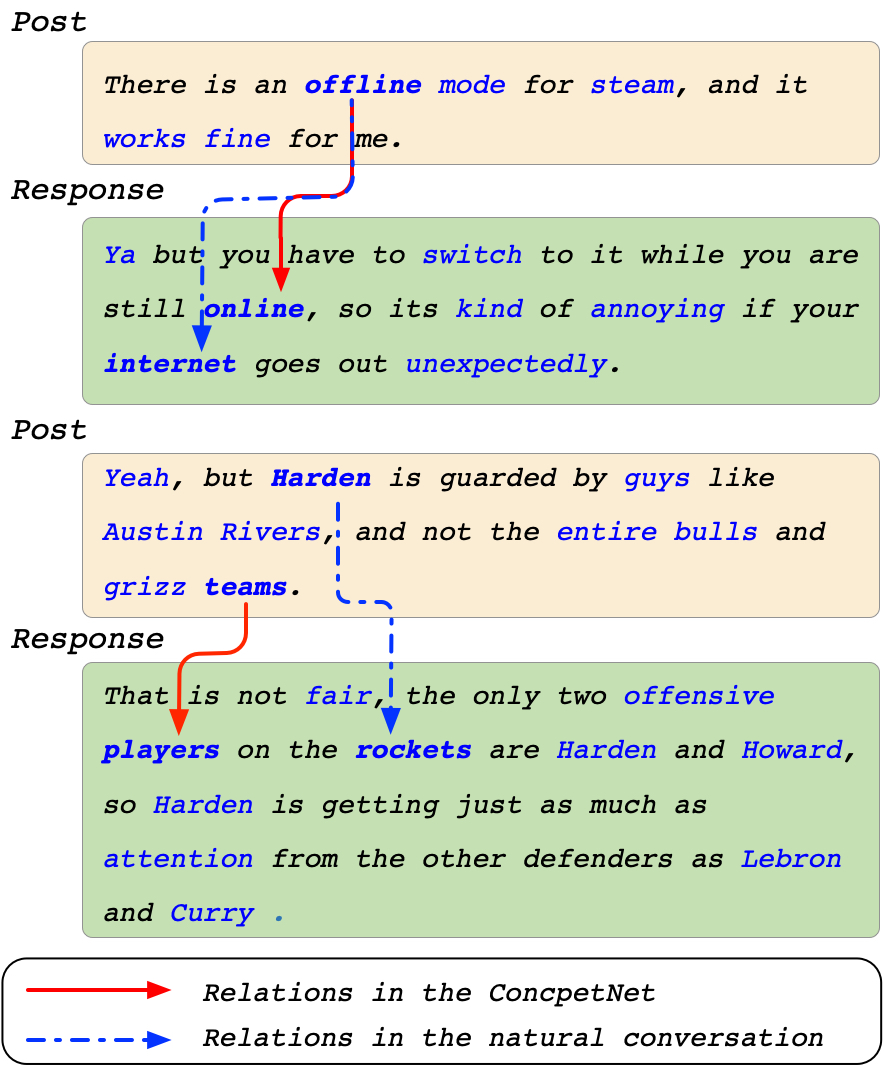}
\caption{Two cases in the Reddit dataset. We use the ConceptNet as the external graph to show concept shifts in the conversation. Nodes are marked in \textcolor{blue}{blue}. Concept relations in the graph and those in the natural conversation are marked with \textcolor{red}{red} solid lines and \textcolor{blue}{blue} dashed lines, respectively.}
\label{fig:example}
\end{figure}
With the rapid development of the natural language generation models \cite{Radford2019language, ZhangSGCBGGLD20, BrownMRSKDNSSAA20} and the increase of the open-domain conversation corpora  \cite{RashkinSLB19, CuiWLZZ20, ZhouZHHZ20, KielaWZDUS18}, the quality of the response generated by the chatbot has been significantly improved. However, there still exist a series of challenges in the generative chatbot \cite{GaoLZBGGD19, HuangZG20}. Most of the time, users can still clearly distinguish between a human talker and a machine chatbot. Part of the reason is that the human is good at naturally switching the topics across the utterances, while the chatbot is relatively dull and tends to keep the topic still \cite{fang2018sounding} or throw an unexpected topic \cite{WangHXSN18, TangZXLXH19}.

As topic flows in the natural conversation could be observed as the shifts of the mentioned concepts across utterances, \newcite{ZhangLXL20} employ the \textbf{ConceptNet} \cite{SpeerCH17} as the external knowledge graph and suggest that the graph provides relation-based one-hop and two-hop concepts to help the response generation. 
Their work is established on a restricted logical assumption: people would like to continuously talk on concepts that have commonsense relations to the current concepts in the ConceptNet. 
We argue the assumption is too simple to imitate topic flows in human conversations.
The ConceptNet is a commonsense graph built based on the concepts and their relations in the real world instead of in the natural conversational scenarios. Thus, only introducing the ConceptNet is insufficient for guiding the response generation. Figure \ref{fig:example} presents two instances in the Reddit conversation dataset for further explanation. Nodes and edges in the ConceptNet are marked to show concept shifts in conversations. For some concept relations that are common in the natural conversation, such as from ``\textcolor{blue}{offline}'' to ``\textcolor{blue}{internet}'' and from ``\textcolor{blue}{Harden}'' to ''\textcolor{blue}{rockets}'', there are not corresponding edges in the ConceptNet.
Therefore, only exploiting knowledge information in the ConceptNet could not cover topic flows in the natural conversation comprehensively.




To address the issue, we propose to reconstruct an enhanced graph that consists of concept relations in both the commonsense knowledge graph and the natural conversation. 
Specifically, we extract new concepts as nodes and the high-frequency concurrence between concepts as edges from the conversation corpora. We then add these new nodes and new edges to the ConceptNet to reconstruct the enhanced graph, which is used at the training and inference procedure for providing hints for the target response.
Besides, we design a novel, powerful, and fast Transformer architecture named \textbf{Edge-Transformer} to encode the enhanced graph, replacing the Graph Neural Networks(GNN).



We conduct experiments on the Reddit conversation dataset. The experimental results show our method outperforms strong baselines and achieves new state-of-the-art performances on many metrics.
We further conduct a series of analysis experiments, which results individually indicate the effectiveness of our proposed enhanced graph and the Edge-Transformer architecture.
Our contributions could be summarized as follows: 
\begin{itemize}
    \item To bridge the gap between concept relations in the external knowledge graph and those in the natural conversation, we reconstruct an enhanced graph with new nodes and edges extracted from the conversation corpora.
    \item We design a novel, powerful and fast architecture named \textbf{Edge-Transformer} that replaces the traditional GNN architecture to encode the enhanced graph. 
    \item Plenty of experiments verify the effectiveness of our method and the importance of concept relations in the conversation corpora. Our method achieves a new state-of-the-art performance on the Reddit conversation dataset.
\end{itemize}

\section{Related Work}

\begin{figure*}[ht]
\centering
\includegraphics[width=\textwidth]{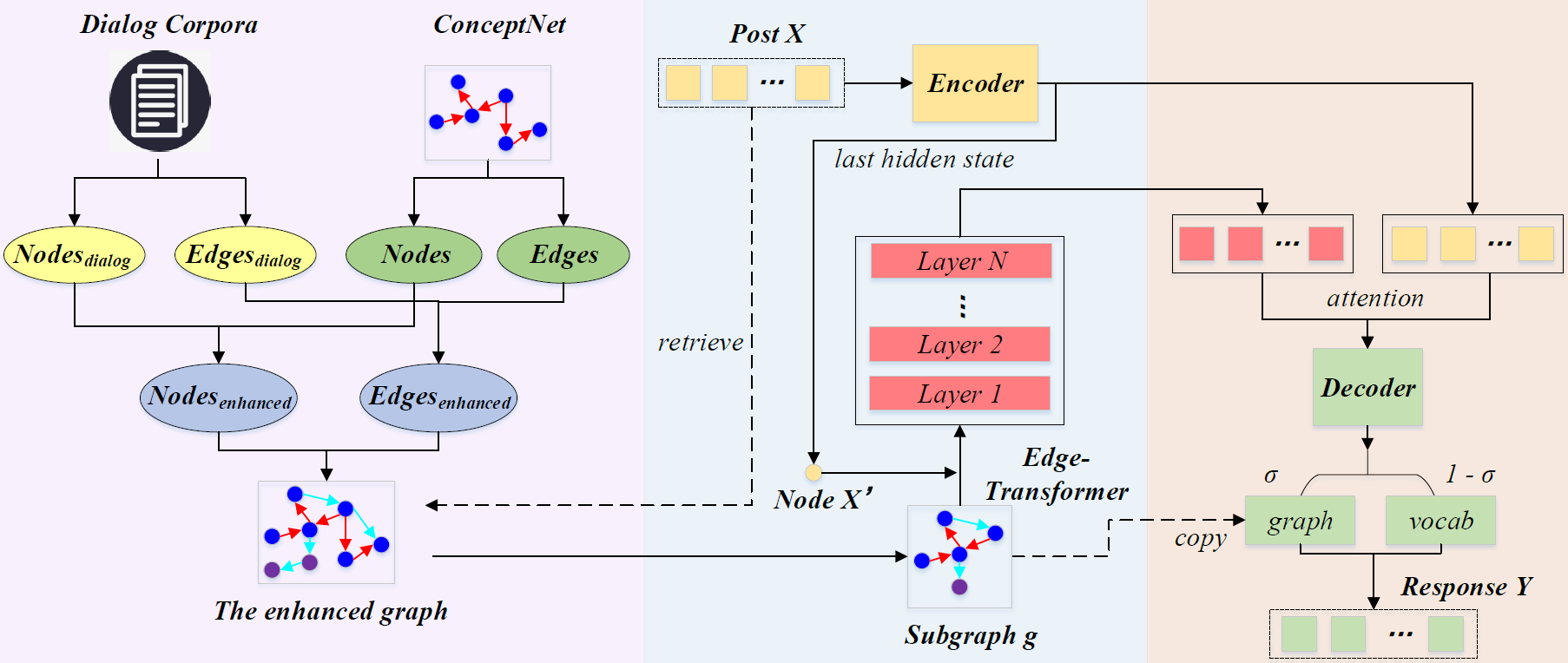}
\caption{The pipeline of our method. Firstly, we extract nodes and edges from the dialog corpora, then merge them with nodes and edges in the ConceptNet, to reconstruct the enhanced graph. Secondly, we retrieve a subgraph $g$ according to the post $X$. We also get a special node $X'$ by the last hidden state of the encoder. $g$ and $X'$ are then encoded by the Edge-Transformer architecture we design. Thirdly, the output of the Edge-Transformer and the output of the encoder are fed into the decoder for response generation. We implement the attention mechanism and the copy mechanism to help generation.}
\label{fig2}
\end{figure*}

The end-to-end generative chatbot \cite{SutskeverVL14} achieves better performance in recent years due to more powerful model architectures\cite{Radford2019language, ZhangSGCBGGLD20, BrownMRSKDNSSAA20} and larger conversation corpora \cite{ZhengCHLZ19, CuiWLZZ20}. However, there also exist a series of challenges in the response generation \cite{HuangZG20}, such as off-topic and uninformative responses\cite{GaoLZBGGD19}. Based on the fact that the natural conversation depends on human knowledge, many works introduce various knowledge, such as background documents \cite{ZhouPB18, GhazvininejadBC18}, commonsense knowledge base\cite{ZhuMZZPY17}, external knowledge graphs\cite{MoonSKS19} for constructing human-like chatbot.

\newcite{ZhouHZZL18} exploit concept relations in the ConceptNet, to imitate concept shifts in human conversation. For a post, they retrieve a subgraph from the ConceptNet, which consists of 0-hop nodes, 1-hop nodes, and edges between these nodes. The subgraph is encoded by the GNN architecture and then introduced to the response generation model. Following this work, \newcite{ZhangLXL20} add 2-hop nodes to the subgraph to cover human concept shifts more comprehensively. We also exploit the ConceptNet, but we argue that only utilizing knowledge information in the ConceptNet is not sufficient because of the gap between concept relations in the commonsense knowledge graph and those in the natural conversation. Thus, we propose to enhance the ConceptNet with the conversation corpora. What's more, to encode the enhanced graph, we design a novel, powerful and fast architecture named Edge-Transformer that replaces the traditional GNN architecture.

There also exist works that directly construct the conversation graph from the real conversation corpora for improving the response generation \cite{TangZXLXH19, XuWNWCL20}. The conversation graph only contains knowledge in the corpora, so its quality is affected by the corpora. In contrast, our enhanced graph is of higher quality because it is built based on the ConceptNet and contains human commonsense knowledge.




\section{Method}

We present our method in this section. We first introduce the overview of our method, then describe three steps of the pipeline in detail.



\subsection{Overview}

Given a conversation corpus $D = \{(X, Y)\}$ where $(X, Y)$ is a dialogue pair in the corpus, we aim to generate the response $Y$ based on the post $X$. With an external knowledge graph $G = (V, E)$ where $V$ and $E$ are nodes and edges in the graph, our task could be formulated as generating best hypothesis $Y'$ which maximizes the following probability:
\begin{align}
    Y' = argmaxP(Y|X, G)
\end{align}

We propose a three-stage method for the task, and the pipeline is presented in Figure \ref{fig2}. Firstly, to bridge the gap between concept relations in the ConceptNet and those in the human conversation, we enhance the ConceptNet with the conversation corpora and reconstruct an enhanced graph $G_e$. Specifically, we extract new nodes and new edges from the conversation corpora $D$, then add them into the ConceptNet $G$. Secondly, since introducing the whole graph to the generation process is unpractical and unnecessary, we retrieve a subgraph $g$ from $G_e$ according to the post $X$. We then design a novel, powerful and fast architecture named Edge-Transformer to replace the traditional GNN architecture. The subgraph $g$ and a special node $X'$ are fed to the Edge-Transformer architecture. Thirdly, to ensure the generation process is guided by knowledge information in $g$, we apply the attention mechanism and the copy mechanism to the classical encoder-decoder framework so that decoder could give responses based on the subgraph $g$, the output of the Edge-Transformer architecture and the output of the encoder.

\subsection{Reconstruct the Enhance Graph}


For some concepts not in the ConceptNet,they are important and frequent in the conversation corpora. Therefore, we add them to the ConceptNet as new nodes so that the enhanced ConceptNet could cover more concepts. We set a frequency interval based on the word frequency of nodes in the ConceptNet. To ensure the extracted nodes have rich semantic information, we choose nouns in the interval as new nodes.


For some concept relations common in the natural conversation, there are not corresponding edges in the ConceptNet. Therefore, we extract some new edges from the conversation corpus and add them to the ConceptNet, to ensure that the enhanced graph could cover concept shifts more comprehensively. We run the GIZA++ alignment tool \footnote{http://www.statmt.org/moses/giza/GIZA++.html} \cite{och03:asc} to align concepts. For a pair of concepts with high alignment probability, we add a new edge between them in the ConceptNet, and the edge has a new category: ``DialogFlowTo''. More details are given in the Appendix we provide.


\subsection{The Edge-Transformer Architecture}



\begin{figure}[ht]
\centering
\includegraphics[width=\columnwidth]{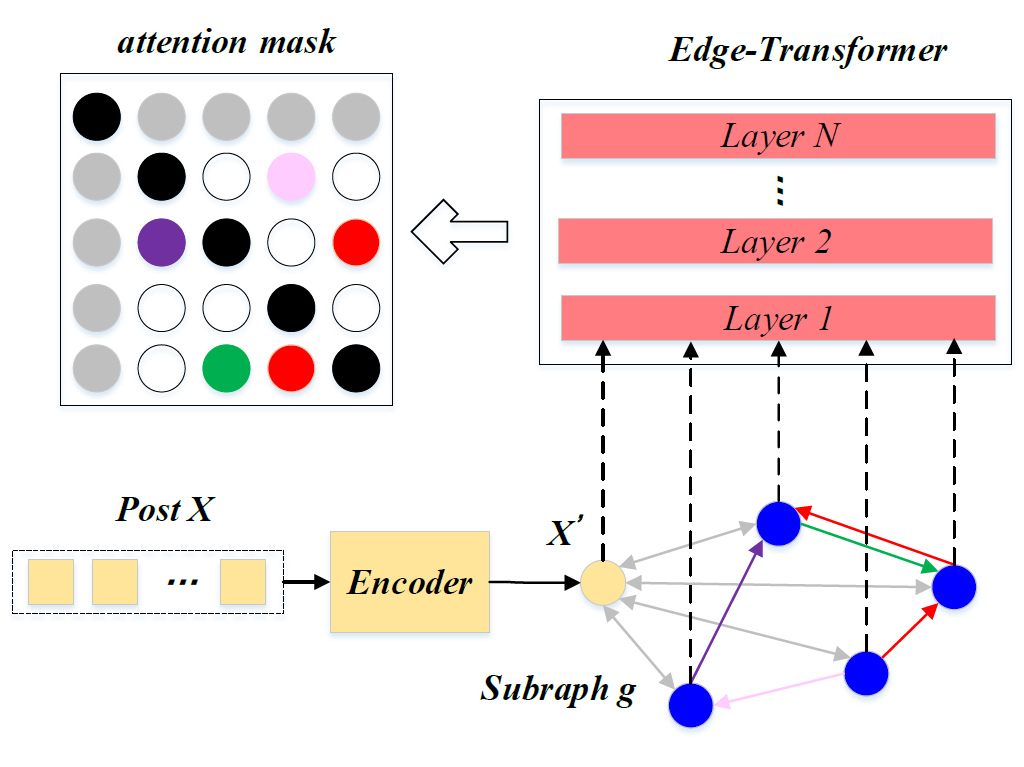}
\caption{Our proposed Edge-Transformer architecture. We show how we use the architecture to encode the subgraph $g$ and the special node $X'$. Attention mask corresponds to edges in the graph structure.}
\label{fig3}
\end{figure}

We make three novel improvements in the vanilla Transformer architecture and propose our Edge-Transformer architecture, which is presented in Figure \ref{fig3}. Firstly, to model the interaction between the post $X$ and the subgraph $g$, we get a special node $X'$ by encoding $X$. $X'$ is added to $g$ and connected to all nodes in $g$. Secondly, the vanilla Transformer architecture can only be used to model the directed complete graph, because each node can obtain information from all other nodes through the attention mechanism. To address the problem, we alter the attention mask in the architecture. Specifically, if there is no edge $(a, b)$ from node $a$ to node $b$ in the graph, we will mask the attention from $b$ to $a$. In this way, the target node could only get information from its source nodes and the architecture could model any directed graph. Thirdly, the vanilla Transformer architecture could not model edge type information in the graph, while there are various edges in the enhanced ConceptNet. To address the problem, we introduce edges information to the forward calculation process of the architecture, as follows:
\begin{align}
    h_{p}^{(l+1)} &= FFN(h_{p}^{(l)} + u_{p}^{(l)})\\
    u_{p}^{(l)} &= \sum\limits_{q \in S(p)} a_{p, q}^{(l)} V^l(h_{q}^{(l)})\\
    a_{p, q}^{(l)} &= Q^{(l)}(h_{p}^{(l)})K^{(l)}(h_{q}^{(l)})^T + R^{(l)}(e_{q, p})
\end{align}

Where $h_{p}^{(l)}$ is the vector of node $p$ in the $l$ layer, and $u_p^{(l)}$ is information from source nodes of $p$ in the $l$ layer. $S(p)$ is source nodes set of $p$, and $a_{p,q}^{(l)}$ is the attention weight. $Q^{(l)}, K^{(l)}, V^{(l)}, R^{(l)}$ are different FFN networks in the $l$ layer, and $e_{q, p}$ is the type of edge $(q,p)$ \footnote{For edges from a node to itself, we give them a new category: ``SelfTO''. For edges from and to $X'$, we give them two new categories: ``FromText'' and ``ToText''.}. 


\subsection{Response Generation}

For the post $X$, the subgraph $g$ contains concepts often thought of in the natural conversation scenario. We implement the attention mechanism on the output of the Edge-Transformer architecture, to ensure the decoder could focus on proper concepts during the generation process. When generating $t$-th response token, the decoder state $s_t$ is updated as follows:
\begin{align}
    s_t = f_{dec}(s_{t-1}, y_{t-1}, c_{t-1}^{text}, c_{t-1}^{graph})
\end{align}

Where $y_{t-1}$ is the token generated in the last step. $c_{t-1}^{text}$ and $c_{t-1}^{graph}$ are outputs of the attention mechanism from the post and the subgraph, respectively. $f_{dec}$ are the updating function of the decoder.

\begin{table*}[ht]
\small
\centering
\begin{tabular}{c|c|c|c|c|c|c|c|c|c}
\hline
\multirow{2}{*}{\textbf{graph}} & \multirow{2}{*}{\textbf{nodes}} & \multirow{2}{*}{\textbf{edges}} & \multirow{2}{*}{\textbf{response nodes}} & \multicolumn{2}{c|}{\textbf{0-hop nodes}} & \multicolumn{2}{c|}{\textbf{1-hop nodes}} & \multicolumn{2}{c}{\textbf{2-hop nodes}} \\ \cline{5-10} 
        &   &   &                        & amount         & golden        & amount          & golden       & amount         & golden        \\ \hline
$G$            & 21471     & 120850     & 5.691     & 5.8129         & 0.5998              & 90.5138         & 1.2064             & 99.7706        & 0.8823              \\ 
$G_e$            & 21754     & 218478     & 6.192    & 6.3223         & 0.6352              & 100.6227        & 1.4114             & 99.7706        & 0.8823              \\ \hline
\end{tabular}
\caption{Statistics of graphs coverage on the conversation dataset. Amount and golden are the number of total concepts and concepts appearing in responses, respectively. Obviously, $G_e$ has a higher coverage than $G$.}
\label{tab:graph statistics}
\end{table*}

Humans usually mention concepts related to the current topic during the conversation. To imitate this phenomenon, we implement the copy mechanism so that the decoder could direct copy nodes from the subgraph as output tokens. We design a binary scalar $\sigma$ as a gate to control the generation source: vocabulary or the subgraph. Thus, the generation probability is the sum of probability on these two sources. The calculation process of $t$-th response token could be formulated as follows:
\begin{align}
    \sigma &= FFN(s_t) \\
    p_t &= (1 - \sigma) p_t^{vocab} + \sigma p_t^{copy} \\
    p_t^{vocab} &= FFN(s_t) \\
    p_t^{copy} &= FFN(a_t)
\end{align}
Where $p_t$, $p_t^{vocab}$ and $p_t^{copy}$ are total prob, prob from vocabulary and prob from the subgraph, respectively. And $a_t$ is the attention weight on the output of the Edge-Transformer architecture. We use the cross-entropy function as the loss to train our model. And the loss of our method contains three parts: the generation loss, the copy loss, and the gate loss, as follows:
\begin{align}
    \mathcal{L} = \mathcal{L}_{gen} + \mathcal{L}_{copy} + \mathcal{L}_{gate}
\end{align}

\section{Experiment}

\subsection{Dataset}

We conduct our experiments on Reddit conversation dataset \cite{ZhouYHZXZ18}. The dataset is a single turn open-domain dialogue dataset, and all utterances are collected from Reddit. The dataset contains 3,384,160 training pairs and 10,000 testing pairs. We use the preprocessed ConceptNet as the external knowledge graph \cite{SpeerCH17},  which includes 21,471 nodes and 120,850 edges. And there are 44 types of edges in the graph.

\subsection{Baselines} 

We follow \citet{ZhangLXL20} and use three groups of models as baselines. We list them here:

\begin{itemize}
    \item \textbf{Standard seq2seq model}\cite{SutskeverVL14}. The model is based on the classical encoder-decoder framework. The encoder and the decoder are RNN architectures. 
    \item \textbf{Knowledge enhanced seq2seq models}: MemNet\cite{GhazvininejadBC18}, CopyNet\cite{ZhuMZZPY17}, CCM\cite{ZhouYHZXZ18} and ConceptFlow\cite{ZhangLXL20}. These models introduce knowledge information into the generation process.
    \item \textbf{Pretraind Models:} GPT-2 lang\cite{ZhangLXL20}, GPT-2 conv\cite{ZhangLXL20}, DialoGPT\cite{ZhangSGCBGGLD20}. These models have a large number of parameters and have been pretrained on large corpus. GPT-2 lang and GPT-2 conv are built based on GPT-2\cite{Radford2019language}.
\end{itemize}

For seq2seq, MemNet, CopyNet, CCM, GPT-2 lang and GPT-2 conv, we directly use results in ConceptFlow paper \cite{ZhangLXL20}. For ConceptFlow, we run their public codes\footnote{https://github.com/thunlp/ConceptFlow.}. For DialoGPT, we finetune it on the dataset \footnote{https://huggingface.co/microsoft/DialoGPT-medium}.
\subsection{Evaluation Metrics}
\subsection{Automation Evaluation}

\begin{table*}[ht]
\small
\centering
\begin{tabular}{c|c|c|c|c|c|c|c|c|c|c}
\hline
\textbf{model} & \textbf{Bleu-3} & \textbf{Bleu-4} & \textbf{Nist-3} & \textbf{Nist-4} & \textbf{Rouge-1} & \textbf{Rouge-2} & \textbf{Rouge-L} & \textbf{meteor} & \textbf{PPL}   & \textbf{Ent-4}   \\ \hline \hline
seq2seq        & 0.0226          & 0.0098          & 1.1056          & 1.1069          & 0.1441           & 0.0189           & 0.1146           & 0.0611          & 48.79          & 7.6650           \\ \hline
MemNet         & 0.0246          & 0.0112          & 1.1960          & 1.1977          & 0.1523           & 0.0215           & 0.1213           & 0.0632          & 47.38          & 8.4180           \\ 
CopyNet        & 0.0226          & 0.0106          & 1.0770          & 1.0788          & 0.1472           & 0.0211           & 0.1153           & 0.0610          & 43.28          & 8.4220           \\ 
CCM            & 0.0192          & 0.0084          & 0.9082          & 0.9095          & 0.1538           & 0.0211           & 0.1245           & 0.0630          & 42.91          & 7.8470           \\ 
ConceptFlow    & 0.0495          & 0.0239          & 1.8838          & 1.8896          & 0.2241           & 0.0457           & 0.2032           & 0.0956          & 29.44          & 10.2390          \\ \hline
GPT-2(lang)    & 0.0162          & 0.0162          & 1.0840          & 1.0844          & 0.1321           & 0.0117           & 0.1046           & 0.0637          & 29.08*          & \textbf{11.6500}          \\ 
GPT-2(conv)    & 0.0262          & 0.0124          & 1.1745          & 1.1763          & 0.1514           & 0.0222           & 0.1212           & 0.0629          & 24.55*          & 8.5460           \\ 
DialoGPT       & 0.0189          & 0.0095          & 0.9986          & 0.9993          & 0.0985           & 0.0117           & 0.0971           & 0.0546          & \textbf{18.65*}          & 9.8163           \\ \hline

Ours           & \textbf{0.0644} & \textbf{0.0331} & \textbf{2.2573} & \textbf{2.2661} & \textbf{0.2592}  & \textbf{0.0601}  & \textbf{0.2340}  & \textbf{0.1091} & 25.98 & 10.8173 \\ \hline
\end{tabular}
\caption{Evaluation results on automatic metrics. We bold the best scores on each metric.  The PPL scores of pretrained models are not comparable because of different tokenization. The results indicate that our method gets the highest scores on most metrics. More results are in the Appendix we provide.}
\label{tab:automation}
\end{table*}

We use following metrics for evaluation:

\begin{itemize}
    \item \textbf{Perplexity} \cite{SerbanSBCP16}: Perplexity measures the fluency of the responses.
    \item \textbf{Bleu \cite{ChenC14}, Nist \cite{DoddingtonG02}, ROUGE\cite{LinY04}} : These metrics measure the overlap between the generated responses and the ground truth.
    \item \textbf{Meteor} \cite{AlonAA07}: Meteor measure the relevance between the generated responses and the ground truth.
    \item \textbf{Entropy} \cite{ZhangGGGLBD18}: Entropy measures the diversity of generated responses.
\end{itemize}

We implement the above metrics based on the code of \citet{Galley18} \footnote{https://github.com/DSTC-MSR-NLP/DSTC7-End-to-End-Conversation-Modeling}.

\subsection{Implementation Details}

Since ConceptFlow \cite{ZhangLXL20} has processed the Reddit conversation dataset with the ConceptNet, we rebuild the dataset based on their data, and details could be found in the Appendix we provide.
Table \ref{tab:graph statistics} presents the coverage of the ConceptNet and our enhanced graph on the Reddit conversation dataset. 


For our model, we use two-layer GRUs \cite{ChoMGBBSB14} as the encoder and the decoder. We set the layers of our Edge-Transformer architecture to 3. We choose Adam as the optimizer, and the batch size, learning rate, max gradients norm, dropout are set to 30, 1e-4, 5, 0.2, respectively. We use TransE embedding \cite{BordesUGWY13} and Glove embedding \cite{PenningtonSM14} to initialize the embedding of concepts and words, respectively. We train our method on 8 V100 GPUs, and it takes about 1.5 hours to train an epoch. Our codes are presented in the supplementary materials.

\section{Evaluation}

\begin{table}[ht]
\small
\centering
\begin{tabular}{c|c|c|c}
\hline
\multirow{2}{*}{\textbf{}}        & \multicolumn{3}{c}{\textbf{Fluency}}                \\ \cline{2-4} 
                                       & \textbf{Average} & \textbf{Best @1} & \textbf{kappa} \\ \hline
\textbf{ConceptFlow}                   & 2.2875           & 0.24             & 0.563          \\ 
\textbf{Ours}                          & 2.4325           & 0.30             & 0.603          \\ 
\textbf{Golden}                        & \textbf{2.6975}           & \textbf{0.69}             & \textbf{0.665}          \\ \hline
\multicolumn{1}{l|}{\multirow{2}{*}{}} & \multicolumn{3}{c}{\textbf{Appropriateness}}        \\ \cline{2-4} 
\multicolumn{1}{l|}{}                  & \textbf{Average} & \textbf{Best @1} & \textbf{kappa} \\ \hline
\textbf{ConceptFlow}                   & 1.6200           & 0.12             & 0.480       \\ 
\textbf{Ours}                          & 1.6850           & 0.16             & 0.563          \\ 
\textbf{Golden}                        & \textbf{2.3275}           & \textbf{0.81}             & \textbf{0.603}          \\ \hline
\end{tabular}
\caption{Evaluation results by human annotators. We also present Fleiss’ Kappa in the table. Kappa values range from 0.4 to 0.6, indicating fair agreement.}
\label{tab:human}
\end{table}

The evaluation results are shown in Table \ref{tab:automation}. Except pretrain models, our method achieves the lowest PPL score, indicating that the responses generated by our model are more fluent. Bleu, Nist, Rouge, and meteor measure the relevance of generated responses and ground truth responses on different aspects. Our method outperforms all baselines by large margins on these metrics, demonstrating the responses generated by our method are more on-topic.

For entropy, our method gets the second-highest score, just lower than GPT-2. It proves that our proposed method could generate diverse responses. 
GPT-lang gets the highest diversity score, but it gets the lowest scores in most relevance metrics like Nist and Rouge. In comparison, our method has a good balance in relevance and diversity.


\subsection{Human Evaluation}

\begin{table*}[ht]
\small
\centering
\begin{tabular}{c|c|c|c|c|c|c|c|c}
\hline
\textbf{model}                               & \textbf{Bleu-3} & \textbf{Bleu-4} & \textbf{Nist-3} & \textbf{Nist-4} & \textbf{Rouge-L} & \textbf{meteor} & \textbf{PPL} & \textbf{Ent-4} \\ \hline \hline
Ours($G_e$ + edge-Transformer) & 0.0644          & 0.0331          & 2.2573          & 2.2661          & 0.2340           & 0.1091          & 25.98        & 10.8173        \\ 
$G$ + edge-Transformer       & 0.0615          & 0.0319          & 2.1448          & 2.1541          & 0.2307           & 0.1055          & 26.40        & 10.7081        \\ 
$G_e$ + GRAFT-Net   & 0.0529  &    0.0267       &    1.9270       &    1.9340       &      0.2115 
     &     0.0976       &   27.81        &     10.4316      \\ 
ConceptFlow($G$ + GRAFT-Net)  & 0.0493          & 0.0246          & 1.8265          & 1.8329          & 0.1888           & 0.0942          & 29.90        & 10.2700        \\ \hline
\end{tabular}
\caption{Evaluation results of models with different combinations of graphs and graph encoding architectures. The results show that $G_e$ outperforms $G$ and the Edge-Transformer outperforms the GRAFT-Net.}
\label{tab:control}
\end{table*}

\begin{table*}[ht]
\small
\centering
\begin{tabular}{c|c|c|c|c|c|c|c|c}
\hline
\textbf{model}                    & \textbf{Bleu-3} & \textbf{Bleu-4} & \textbf{Nist-3} & \textbf{Nist-4} & \textbf{Rouge-L} & \textbf{meteor} & \textbf{PPL} & \textbf{Ent-4} \\ \hline \hline
enhanced graph                    & 0.0644          & 0.0331          & 2.2573          & 2.2661          & 0.2340           & 0.1091          & 25.98        & 10.8173        \\ 
- edges in bottom 20\%  & 0.0634          & 0.0328          & 2.2102          & 2.2194          & 0.2322           & 0.1070          & 27.17        & 10.7391        \\ 
- edges in bottom 50\% & 0.0502          & 0.0249          & 1.8466          & 1.8528          & 0.2044           & 0.0938          & 30.77        & 10.2637        \\ \hline
\end{tabular}
\caption{Evaluation results after removing edges in the ConceptNet. More results are  in the Appendix.}
\label{tab:effective}
\end{table*}

To further evaluate model performances, we hire four human annotators to judge the quality of generated responses. Annotators are required to score the responses on two aspects: fluency and appropriateness. Fluency evaluates whether a response is fluent or contains any grammar errors, while appropriateness evaluates whether a response is relevant to its post. Specifically, we sample 100 cases for three methods: ConceptFlow, ours, and golden (ground truth responses), and all responses are scored from 1 to 3 on two aspects.


Human evaluation result is shown in Table \ref{tab:human}. Obviously, ground truth responses get the highest average scores. The average scores of our method are higher than the scores of ConceptFlow on both aspects, indicating our method could give more fluent and more relevant responses. And the best @1 ratios of our method are also higher than ConceptFlow, demonstrating that humans are more willing to chat with our chatbot. The results of the automatic evaluation and human evaluation prove the effectiveness of our method. With the enhanced graph and the Edge-Transformer architecture, our method could give responses of higher quality.  Next, we conduct a series of experiments to study the effectiveness of the enhanced graph and the Edge-Transformer architecture individually.

\subsection{Analysis of the Enhanced Graph}
\label{graph analysis}

In this part, we conduct a series of experiments to study the effectiveness of the enhanced graph  $G_e$.

\textbf{The enhanced graph VS the ConceptNet.} Considering that our method utilizes the enhanced graph and the edge-Transformer architecture ($G_e$ + edge-Transformer) while ConcpetFlow \cite{ZhangLXL20} utilizes the original ConceptNet and the GNN-based architecture named GRAFT-Net \cite{SunDZMSC18} ($G$ + GRAFT-Net), we conduct two more models to directly compare $G_e$ and $G$. The first model is built on $G$ + Edge-Transformer, and the second is built on $G_e$ + GRAFT-Net. The result is presented in Table \ref{tab:control}. Obviously, with the same graph encoding architecture, models with $G_e$ achieve better performances on all metrics than models with $G$. The comparison results show that $G_e$ is more helpful to the response generation. And the importance of concept relations from the conversation corpora is also proved.

\textbf{Concept relations from the conversations corpora VS those in the ConceptNet.} Now that we prove concept relations from the conversation corpora are important for the response generation, there is one more question to answer: Is it enough to only exploit concept relations from the conversation corpora? In other words, is the external commonsense knowledge graph such as the ConceptNet unnecessary? 
To study the question, we remove some edges in the ConceptNet when reconstructing the enhanced graph, and implementation details are given in the Appendix. The evaluation result is shown in Table \ref{tab:effective}. Our method gets lower scores on all metrics after reducing edges. And reducing more edges results in worse performances. We could infer that concept relations, which are rare in the natural conversations, are also important for guiding topic flows in the response generation process. Thus, knowledge information in both the external graph and the conversation corpora are necessary, and a good way is to merge them like our method.


\begin{table}[ht]
\small
\centering
\begin{tabular}{c|c|c}
\hline
\textbf{model}     & \textbf{parameters} & \textbf{training time/epoch} \\ \hline
Edge-Transformer &     \textbf{34.6M}           &    \textbf{1.5h}                      \\ 
GRAFTGNN         &     35.3M           &    2.5h                      \\ \hline
\end{tabular}
\caption{Computation resources of different graph encoding architectures.}
\label{tab:computation}
\end{table}

\begin{table*}[ht]
\small
\centering
\begin{tabular}{c|c|c|c|c|c|c|c|c|c|c}
\hline
\textbf{model}     & \textbf{Bleu-3} & \textbf{Bleu-4} & \textbf{Nist-3} & \textbf{Nist-4} & \textbf{Rouge-1} & \textbf{Rouge-2} & \textbf{Rouge-L} & \textbf{meteor} & \textbf{PPL} & \textbf{Ent-4} \\ \hline \hline
Ours               & 0.0644          & 0.0331          & 2.2573          & 2.2661          & 0.2592           & 0.0601           & 0.2340           & 0.1091          & 25.98        & 10.8173        \\ \hline
w/o post node      & 0.0595          & 0.0305         & 2.1316           & 2.1402         & 0.2487             & 0.0562           & 0.2237          & 0.1044           & 27.00        & 10.7731        \\
w/o edge mask      & 0.0573          & 0.0290         & 2.0694           & 2.0771         & 0.2442 
    & 0.0538           & 0.2201           & 0.1025          & 26.81      & 10.6822          \\
w/o edge emb       & 0.0589          & 0.0295          & 2.1394          & 2.1472          & 0.2485 
    & 0.0547           & 0.2246           & 0.1050          & 26.46        & 10.6871        \\ \hline
\end{tabular}
\caption{Automation results of ablation models. All ablation models get lower scores than the complete model.}
\label{tab:ablation}
\end{table*}

\begin{table*}[ht]
\small
\centering
\begin{tabular}{c|l}
\hline
\textbf{\#1 post} & \begin{tabular}[c]{@{}l@{}}I \textcolor{blue}{drove} \textcolor{blue}{home} last \textcolor{blue}{night} , saw my \textcolor{blue}{dad} for the first \textcolor{blue}{time} in 6th \textcolor{blue}{months} , and \textcolor{blue}{slept} all \textcolor{blue}{day} \textcolor{blue}{today} . \\ \textcolor{blue}{Woke} up and \textcolor{blue}{poured} myself a \textcolor{blue}{rum} and \textcolor{blue}{coke} then\textcolor{blue}{started} \textcolor{blue}{watching} \textcolor{blue}{archer} . \textcolor{blue}{Today} was a good day .\end{tabular} \\ \hline
DialoGPT          & I 'm glad you had a good day.                                                                                                                                                                                                                \\
ConcpetFlow       & I 'm going to the same \textcolor{blue}{day} . Have a good \textcolor{blue}{night} , \textcolor{blue}{man} .                                                                                                                                                                                      \\
Ours              & I 'm going to \textcolor{blue}{watch} the first \textcolor{magenta}{episode} of \textcolor{blue}{archer} and see if i can get a chance to \textcolor{blue}{sleep} .                                                                                                                                                    \\ \hline
\textbf{\#2 post} & \begin{tabular}[c]{@{}l@{}}What do you mean ? From the \textcolor{blue}{alpha} or from the \textcolor{blue}{beta} ? His \textcolor{blue}{uav} was nerfed like \textcolor{blue}{crazy} in the \textcolor{blue}{beta} , \\ not being able to ping or \textcolor{blue}{find} a \textcolor{blue}{monster} if they 're \textcolor{blue}{sneaking} at all .\end{tabular}                          \\ \hline
DialoGPT          & I mean from the alpha . I'm not sure if it was nerfed in the beta, but i'm pretty sure it was.                                                                                                                                               \\
ConcpetFlow       & You can see the \textcolor{blue}{alpha} of the \textcolor{blue}{alpha} , but it 's a \textcolor{blue}{bug} .                                                                                                                                                                                      \\
Ours              & You can see the \textcolor{magenta}{source} on the \textcolor{magenta}{server} . I think he was just a \textcolor{blue}{bug} .                                                                                                                                                                          \\ \hline
\end{tabular}
\caption{Two cases on the testset. We present responses generated by different models. We mark concepts in the original ConceptNet in \textcolor{blue}{blue} and concepts introduced by the enhanced graph in \textcolor{magenta}{magenta}.}
\label{tab:case}
\end{table*}

\begin{figure}[ht]
\centering
\includegraphics[width=\columnwidth]{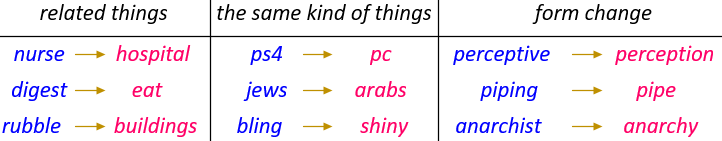}
\caption{Several examples of high-quality concept relations we extracted from the conversation corpora. We classify them into three categories.}
\label{fig4}
\end{figure}

\textbf{Quality evaluation of the extracted edges.} We conduct a human evaluation to verify the quality of the extracted edges. Specifically, we sample 100 extracted edges, and four human annotators are required to judge whether the target concept is relevant to the source concept. On average, 68 concept relations are marked as relevant edges. And there are 47 edges that all four annotators think relevant. We list some of these high-quality edges in Figure \ref{fig4} and classify them into three categories roughly. The first type corresponds a pair of things that have a realistic relationship, such as \textcolor{blue}{``nurse''} works for \textcolor{magenta}{``hospital''}. The second type corresponds a pair of things in the same kind, such as both \textcolor{blue}{``ps4''} and \textcolor{magenta}{``pc''} are electronic devices. The third type corresponds a pairs of concepts with POS relationship, such as \textcolor{magenta}{``perception''} is the noun form of \textcolor{blue}{``perceptive''}. These three categories are consistent with human common sense, proving our method could get various knowledge information from the real conversation corpora.


\subsection{Analysis of the Edge-Transformer Architecture}

In this part, we conduct a series of experiments to study the effectiveness of our proposed Edge-Transformer architecture.

\textbf{The Edge-Transformer VS the GRAFT-Net.} From evaluation results in Table \ref{tab:control}, we could see that with the same graph, models with the Edge-Transformer achieve higher scores on all metrics than models with the GRAFT-Net. The results demonstrate the Edge-Transformer could encode graphs better. We also compare the parameters and training time of two architectures, which results are shown in Table \ref{tab:computation}. Obviously, our architecture contains fewer parameters with high training speed. The above two comparison shows the Edge-Transformer gets better performances than the GRAFT-Net while costing fewer computation resources. 

\begin{table}[ht]
\small
\centering
\begin{tabular}{c|c|c}
\hline
\textbf{words num} & \textbf{concepts in $G_e$} & \textbf{concepts in $G$} \\ \hline
 19.1056          &       2.2001      &        2.0593                   \\ \hline
\end{tabular}
\caption{Concepts num in the generated responses.}
\label{tab:response_nodes}
\end{table}

\textbf{Ablation study of the Edge-Transformer Architecture.} We propose three improvements on vanilla Transformer architecture and build the Edge-Transformer architecture. To study the effectiveness of three improvements, respectively, we build corresponding ablation models, as follows:
\begin{itemize}
    \item \textbf{w/o post node.} We remove the special node $X'$, and there is no interaction between the post $X$ and the subgraph $g$.
    \item \textbf{w/o edge mask.} We remove the edge mask, and the architecture is the vanilla Transformer.
    \item \textbf{w/o edge embed.} We remove the edge embedding in the architecture, and the edge type information is not introduced.
\end{itemize}
The evaluation results of these three ablation models are shown in Table \ref{tab:ablation}. All ablation models get lower scores than the complete model on all metrics. The architecture without edge mask gets the lowest scores, indicating graph structure information in the knowledge graph is vital for the response generation and the vanilla Transformer architecture could not encode graph structures well.
The results also prove the necessity of interaction between the post and the subgraph, and the importance of the edge type information. 

\subsection{Case Study}

To further study the improvement our method brings, we present two cases in Table \ref{tab:case}.  In case 1, DialoGPT and ConcpetFlow generate proper responses, but their responses are not as informative as ours. We could see that our response contains concept  ``\textcolor{magenta}{episode}'' from $G_e$, demonstrating that $G_e$ could bring new concepts to the generated responses. In case 2, it seems that DialoGPT and ConceptFlow don't understand the post and give wrong responses. While our method gives high-quality response that contains concepts ``\textcolor{magenta}{source}'',  ``\textcolor{magenta}{server}'' and ``\textcolor{blue}{bug}'', which are relevant to the post.

Besides, we statistic the concepts in the generated responses on the testset, which is shown in Table \ref{tab:response_nodes}. In generated response, there are 2.2 words in the enhanced graph $G_e$ on average. Compared to the ConceptNet, the enhanced graph indeed introduces new concepts into the responses. The results prove the effectiveness of our method further.

\section{Conclusion}

Because of the gap between the concept relations in the natural conversation and those in the external commonsense knowledge graph, exploiting the knowledge information in the external knowledge graph is not sufficient to guide topic flows in the response generation. We extract conversation knowledge information from the conversation corpus to enhance the ConceptNet. To improve the knowledge-based response generation, we reconstruct an enhanced graph and design a novel architecture named Edge-Transformer to encode the enhanced graph. Plenty of experiments on the Reddit dataset show our method outperforms other strong baselines, achieving new SOTA results.


\bibliography{custom}
\bibliographystyle{acl_natbib}

\clearpage
\appendix

\section{Data Processing}

This part presents some details of data processing in this paper.

\subsection{Extracting New Nodes and New Edges}
\label{appendix:a-1}

We reconstruct an enhanced graph based on the ConceptNet and the conversation corpus. Specifically, we extract a series of new nodes and new edges, then add them to the ConceptNet.

We choose nouns with frequencies in the corpus as new nodes because these nodes are common and vital in the corpus, meanwhile have rich semantic information. We use the NLTK toolkit in python3 for POS tagging \footnote{https://www.nltk.org/}. And we statistic the frequencies of all nodes in the ConceptNet, then regard the top $m$ frequency as the threshold. We set $m$ to 20\% here. Nouns which frequencies higher than the threshold are the extracted new nodes.

\begin{figure}[ht]
\centering
\includegraphics[width=\columnwidth]{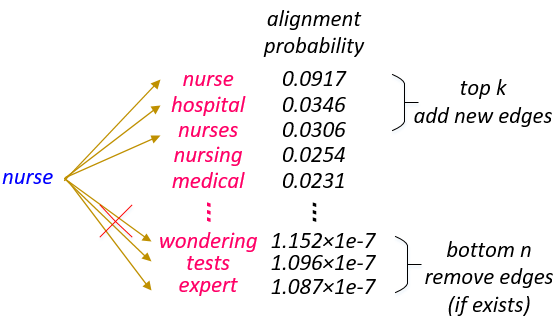}
\caption{An example of the extract edges from the conversation corpus.}
\label{fig5}
\end{figure}

We run the GIZA++ alignment tool \footnote{http://www.statmt.org/moses/giza/GIZA++.html} \cite{och03:asc} to align concepts in the conversation corpus, because the concept alignment reflects concept relations in the natural conversation. For all dialog pairs in the training dataset, we keep concepts and remove other words. The processed dialog pairs are regarded as input data to the GIZA++ tool. The output results are alignment probabilities between concepts, which are utilized to extract new edges. Figure \ref{fig5} presents an example. For the concept ``nurse'', we rank other concepts according to the alignment probabilities. We regard the top $k$ concepts as relevant concepts of ``nurse'' and add edges between ``nurse'' and these concepts in the ConceptNet. We set $k$ to 5 here.  

When we conduct experiments about reducing edges in the ConceptNet, as said in subsection \ref{graph analysis}, we also utilize the alignment probabilities to remove edges. As shown in figure \ref{fig5}, if there exist edges from ``nurse'' to bottom $n$ concepts in the ConceptNet, we will remove these edges. We set $n$ to 20\% and 50\%, respectively.



\subsection{Rebuild the Conversation Dataset}
\label{appendix:a-2}

We conduct our experiments on Reddit conversation dataset \cite{ZhouYHZXZ18}. ConceptFlow \cite{ZhangLXL20} has processed the dataset with the ConceptNet . They get a subgraph for the post $x$, which contains 0-hop, 1-hop, and 2-hop nodes from source nodes $N_x$. Especially, they only keep 100 2-hop nodes in $g$ and remove others.

\begin{algorithm}[h]
\caption{Getting the subgraph $g$}
\label{alg1}
\hspace*{0.02in} {\bf Input:} 
the post $x$ and the enhanced graph $G_e$ \\
\hspace*{0.02in} {\bf Output:} 
the subgraph $g$
\begin{algorithmic}[1]
\STATE Initiate $V_g, E_g = \emptyset$
\STATE Match $x$ and $V_e$ to get source nodes set $V_x$.
\STATE Initiate $V_0 = V_x, V_1 = \emptyset, V_2 = \emptyset$
\FOR{each node $a \in V_0$}
    \STATE Get its neighborhood nodes set $\mathcal{N}_a\subset V_e$.
    \FOR {each node $b \in \mathcal{N}_a$}
        \STATE $E_g = E_g \cup \{(a, b)\}$
        \IF {$b \notin V_0$}
            \STATE $V_1 = V_1 \cup \{b\}$
        \ENDIF
    \ENDFOR
\ENDFOR
\FOR{each node $a \in V_1$}
    \STATE Get its neighborhood nodes set $\mathcal{N}_a\subset V_e$.
    \FOR {each node $b \in \mathcal{N}_a$}
        \IF {$b \notin V_0$ and $b \notin V_1$}
            \IF {$b \in V_{2-base}$}
                \STATE $V_2 = V_2 \cup \{b\}$
                \STATE $E_g = E_g \cup \{(a, b)\}$
            \ENDIF
        \ELSE
            \STATE  $E_g = E_g \cup \{(a, b)\}$
        \ENDIF
    \ENDFOR
\ENDFOR
\STATE $V_g = V_0 \cup V_1 \cup V_2$
\STATE Return $g = (V_g, E_g)$
\end{algorithmic} 
\end{algorithm}

\begin{table*}[ht]
\small
\centering
\begin{tabular}{c|c|c|c|c|c|c|c}
\hline
\textbf{model} & \textbf{Bleu-1} & \textbf{Bleu-2} & \textbf{Nist-1} & \textbf{Nist-2} & \textbf{Dist-1} & \textbf{Dist-2} & \textbf{Concept-PPL}    \\ \hline \hline
seq2seq        & 0.1702          & 0.0579          & 1.0230          & 1.0963          & 0.0123          & 0.0525             & -                  \\ \hline
MemNet         & 0.1741          & 0.0604          & 1.0975          & 1.1847          & 0.0211           & 0.0931           & 46.85            \\ 
CopyNet        & 0.1589          & 0.0549          & 0.9899          & 1.0664          & 0.0233           & 0.0988           & 40.27            \\ 
CCM            & 0.1413          & 0.0484          & 0.8362          & 0.9000          & 0.0146           & 0.0643           & 39.18            \\ 
ConceptFlow    & 0.2495          & 0.1064          & 1.6685          & 1.8531          & 0.0237           & 0.1268           & 26.76            \\ \hline
GPT-2(lang)    & 0.1705          & 0.0486          & 1.0231          & 1.0794          & 0.0325           & \textbf{0.2461}           & -            \\ 
GPT-2(conv)    & 0.1765          & 0.0625          & 1.0734          & 1.1623          & 0.0266           & 0.1218           & -            \\ 
DialoGPT       & 0.1404          & 0.0442          & 0.9195          & 0.9906          & \textbf{0.0632}           & 0.2288           & -            \\ \hline
Ours           & \textbf{0.2872} & \textbf{0.1301} & \textbf{1.9607} & \textbf{2.2123} & 0.0256  & 0.1485  & \textbf{24.68}  \\ \hline
\end{tabular}
\caption{Supplementary evaluation results on automatic metrics. We bold the best scores on each metric.  Some models don't utilize concept information, so Concept\_PPL is not suitable for them.}
\label{tab:sup-automation}
\end{table*}

\begin{table*}[ht]
\small
\centering
\begin{tabular}{c|c|c|c|c|c|c|c|c}
\hline
\textbf{model}                    & \textbf{Bleu-1} & \textbf{Bleu-2} & \textbf{Nist-1} & \textbf{Nist-2} & \textbf{Rouge-1} & \textbf{Rouge-2} & \textbf{Dist-1} & \textbf{Dist-2} \\ \hline \hline
enhanced graph          & 0.2872          & 0.1301          & 1.9607          & 2.2123       & 0.2592           & 0.0601          & 0.0256        & 0.1485        \\ 
- edges in bottom 20\%  & 0.2821          & 0.1276          & 1.9234          & 2.1653       & 0.2591           & 0.0606          & 0.0251        & 0.1463        \\ 
- edges in bottom 50\%  & 0.2455          & 0.1055          & 1.6277          & 1.8144       & 0.2233           & 0.0476          & 0.0238        & 0.1262        \\ \hline
\end{tabular}
\caption{Evaluation results of models when reducing edges in the ConceptNet.}
\label{tab:sup-effective}
\end{table*}

For the fairness of the experiment, we rebuild the conversation dataset with the enhanced graph $G_e$, based on their dataset. For the post $x$, we get a subgraph $g$ in $G_e$, and we present our method in Algorithm \ref{alg1}. Where $V_0, V_1, V_2$ are 0-hop, 1-hop, 2 hop nodes set, respectively. And $V_{2-base}$ is the 2-hop nodes set in ConceptFlow dataset.

\section{Supplementary Evaluation Results}
\label{appendix:b}

This part presents more evaluation results.

\subsection{Supplementary Result for Overall Experiments}
\label{appendix:b-1}


Table \ref{tab:sup-automation} shows supplementary evaluation result of generated responses. We use two new metrics for evaluation. Dist \cite{LiGBGD16} measures the diversity of generated responses, and Concept-PPL\cite{ZhouYHZXZ18} calculates perplexity by considering both entities and words. We could see that our method gets the lowest Concept-PPL, showing the generated responses by our method are most fluent. Our method also achieves the best performances in Bleu and Nist, demonstrating that our method could give the most relevant responses. Pretrained models get the highest diversity scores because of the rich semantic information they get during the pretrain process. Besides these pretrianed models, our method gets the highest diversity scores, showing our responses are the most informative. The supplementary result demonstrates that our method could give responses with higher quality than other baselines, and further confirms the effectiveness of the enhanced graph $G_e$ and the Edge-Transformer architecture.

\subsection{Supplementary Result for Experiments of Reducing Edges}
\label{appendix:b-2}

We present the supplementary evaluation result of models when reducing edges in the ConceptNet in Table \ref{tab:sup-effective}. Obviously, our method gets lower performances on almost all metrics, and removing 50\% edges causes worse results than reducing 20\% edges. The result indicates the edges in the ConceptNet are important and necessary for the response generation. Specifically, we find diversity scores drop a lot when reducing 50\% edges.
The results above further prove that ConceptNet is vital for the generation process.

\end{document}